\documentclass[conference]{IEEEtran}
\IEEEoverridecommandlockouts

\usepackage{cite}
\usepackage{amsmath,amssymb,amsfonts}
\usepackage{algorithmic}
\usepackage{graphicx}
\usepackage{textcomp}
\usepackage{xcolor}
\usepackage{multirow}
\usepackage{tabularx}
\usepackage{svg}
\usepackage{booktabs} 
\usepackage{graphicx} 
\def\BibTeX{{\rm B\kern-.05em{\sc i\kern-.025em b}\kern-.08em
    T\kern-.1667em\lower.7ex\hbox{E}\kern-.125emX}}

\usepackage{acro}
\usepackage{soul}
\usepackage{microtype}
\usepackage[normalem]{ulem}
\usepackage{acro}
\usepackage{combelow}
\DeclareAcronym{mattnet}{
	short = {\scshape MattNet} ,
	long  = \textbf{m}ultimodal \textbf{att}ention \textbf{net}work,
	alt = \textbf{M}ultimodal \textbf{Att}ention \textbf{Net}work
}
\DeclareAcronym{AI}{
	short = AI,
	long  = artificial intelligence,
	alt = Artificial Intelligence
}
\DeclareAcronym{LLM}{
	short = LLM,
	long  = large language models,
	alt = Large Language Models
}
\DeclareAcronym{ML}{
	short = ML,
	long  = machine learning,
	alt = Machine Learning
}
\DeclareAcronym{DL}{
	short = DL,
	long  = deep learning,
	alt = Deep Learning
}
\DeclareAcronym{AE}{
	short = AE ,
	long  = autoencoder,
	alt = Autoencoder
}
\DeclareAcronym{CAE}{
	short = CAE,
	long  = correspondence autoencoder,
	alt = Correspondence Autoencoder
}
\DeclareAcronym{CPC}{
	short = CPC,
	long  = contrastive predictive coding,
	alt = Contrastive Predictive Coding
}
\DeclareAcronym{ME}{
	short = ME,
	long  = mutual exclusivity,
	alt = Mutual Exclusivity
}
\DeclareAcronym{VGS}{
	short = VGS,
	long  = visually grounded speech models
}
\DeclareAcronym{CELL}{
	short = CELL,
	long  = Cross-channel Early Lexical Learning,
	alt = Cross-channel Early Lexical Learning
}
\DeclareAcronym{VQ}{
	short = VQ,
	long  = vector quantised,
	alt = Vector Quantised
}
\DeclareAcronym{CLIP}{
	short = CLIP,
	long = Contrastive Language-Image Pre-training,
	alt = Contrastive Language-Image Pre-training
}
\DeclareAcronym{CNN}{
	short = CNN,
	long = Convolutional Neural Network,
	alt = convolutional neural network
}
\DeclareAcronym{vpkl}{
	short = VPKL,
	long = visually prompted keyword localisation
}
\DeclareAcronym{BOW}{
	short = BoW,
	long = bag-of-words
}

\DeclareAcronym{UN}{
	short = UN,
	long = United Nations
}

\newcommand{\locAtt}{{\small \scshape Loc\-AttNet}}

\newcommand{\qbert}{{\small QbERT}}
\newcommand{\boy}{\d{o}m\d{o}k\`{u}nrin}
\newcommand{\ocean}{\`{o}kun}
\newcommand{\pool}{od\`{o} ad\'{a}g\'{u}n}
\newcommand{\yoruba}{Yor\`{u}b\'{a}}

\newcommand{\image}[1]{{\scshape #1}}
\newcommand{\word}[1]{``#1''}

\usepackage{xcolor}
\definecolor{mycolor}{HTML}{008000}
\definecolor{indiagreen}{HTML}{138808}
\definecolor{papaya}{HTML}{EE892F}
\definecolor{mygreen}{HTML}{008000}
\definecolor{mypurple}{HTML}{9966CC}
\definecolor{mygray}{HTML}{696969}
\definecolor{myblue}{HTML}{5D8AA8}

\newcommand{\smallma}{{\footnotesize \scshape Loc\-AttNet}}

\usepackage{pifont}

\usepackage{booktabs}
\usepackage{multirow}
\usepackage{tabularx}

\newcolumntype{C}{>{\centering\arraybackslash}X}
\newcolumntype{L}{>{\raggedright\arraybackslash}X}
\newcolumntype{R}{>{\raggedleft\arraybackslash}X}
\newcolumntype{P}[1]{>{\raggedright\arraybackslash}p{#1}}

\usepackage[prependcaption,textsize=scriptsize]{todonotes}
\begin{document}

\title{Improved visually prompted keyword localisation in real low-resource settings%
\thanks{This work was supported in part by a grant of the Ministry of Research, Innovation and Digitization, CNCS-UEFISCDI, project number PN-IV-P2-2.1-TE-2023-1632, within PNCDI IV.}}

\author{
\IEEEauthorblockN{Leanne Nortje}
\IEEEauthorblockA{\textit{Electrical and Electronic Engineering}\\
\textit{Stellenbosch University} \\
Stellenbosch, South Africa \\
nortjeleanne@gmail.com}\\
\IEEEauthorblockN{Gabriel Pîrlogeanu}
\IEEEauthorblockA{\textit{National University of Science and Technology}\\
\textit{\textsc{Politehnica} Bucharest} \\
Bucharest, Romania \\
gabriel.pirlogeanu@upb.ro}\\
\and
\IEEEauthorblockN{Dan Oneata}
\IEEEauthorblockA{\textit{National University of Science and Technology}\\
\textit{\textsc{Politehnica} Bucharest} \\
Bucharest, Romania \\
dan.oneata@gmail.com}\\
\IEEEauthorblockN{Herman Kamper}
\IEEEauthorblockA{\textit{Electrical and Electronic Engineering}\\
\textit{Stellenbosch University} \\
Stellenbosch, South Africa \\
kamperh@sun.ac.za}\\
}

\maketitle

\begin{abstract}
    Given an image query, the goal in visually prompted keyword localisation (VPKL) is to find occurrences of the depicted word in a speech collection.
    This can be useful when transcriptions are not available for a low-resource language (e.g.\ if it is unwritten).
    Previous work showed that VPKL can be performed with a visually grounded speech model trained on paired images and unlabelled speech.
    But all experiments were done on English.
    Moreover, transcriptions were used to get positive and negative pairs for the contrastive loss.
    This paper introduces a few-shot learning scheme to mine pairs automatically without transcriptions.
    On English, this results in only a small drop in performance.
    We also -- for the first time -- consider VPKL on a real low-resource language, \yoruba.
    While scores are reasonable, here we see a bigger drop in performance compared to using ground truth pairs because the mining is less accurate in~\yoruba.%
\end{abstract}


\section{Introduction}

\begin{figure}[!b]
	\begin{minipage}[b]{0.99\linewidth}
		\centering
		\centerline{\includegraphics[width=0.9\linewidth]{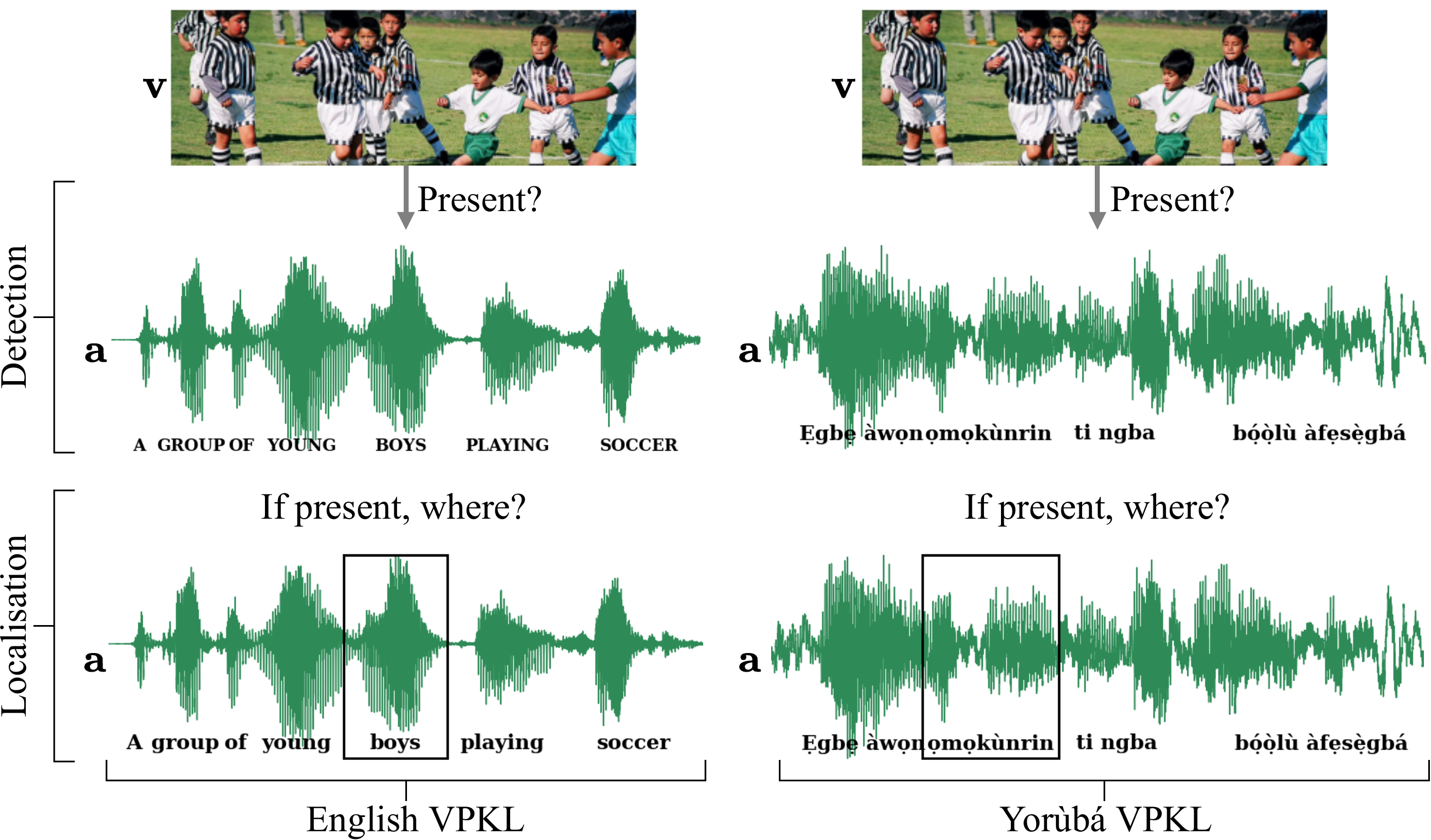}}
	\end{minipage}
    \vspace*{-7pt}
	\caption{The goal in 
		visually prompted keyword localisation is to detect and locate a given query keyword (given as an image) within spoken utterances.
        On the right, the \yoruba\ word for \word{boys} is \word{\boy}. 
  }
	\label{fig:vpkl}
\end{figure}

Developing applications that can search through speech data is challenging in low-resource languages where transcriptions are difficult or impossible to collect.
One line of research has been looking at visually grounded speech models to address this~\cite{olaleye_yfacc_2023}.
These models learn from paired images and unlabelled spoken captions and can therefore be trained without transcriptions~\cite{chrupala_representations_2017, chrupala_visually_2022, scharenborg_speech_2020, scholten_learning_2021, peng_fast-slow_2022, harwath_jointly_2018}. 
One way to perform speech search with such a model is to provide an image query depicting a word of interest.
Formally, the task of visually prompted keyword localisation (VPKL) involves detecting whether an image query~-- which depicts a keyword -- occurs in a spoken utterance, and if so, where it occurs~\cite{nortje_towards_2023}.
An English example is shown on the left in Fig.~\ref{fig:vpkl}.

Previous work~\cite{nortje_towards_2023} formalised the VPKL task and showed that it is possible at a reasonable level with a visually grounded speech model.
However, there were two major shortcomings.
First, all experiments were carried out on English datasets, treating it as an artificial low-resource language.
Second -- and more importantly -- English transcriptions were used to obtain positive and negative pairs for the contrastive loss used in the visually grounded model.
This reliance on transcriptions severely limits the applicability of the approach to a real low-resource setting.
In this paper, we address these shortcomings by performing experiments on \yoruba, a real low-resource language spoken by 44M people in Nigeria. 
A \yoruba\ example is given in the right part of Fig.~\ref{fig:vpkl}.
We also adapt the original approach to work without using transcriptions, making it usable in the low-resource case.

Concretely,
we turn to few-shot learning to mine training pairs~\cite{nortje_visually_2023-1}.
We use a support set that contains a small number of isolated spoken examples of the keywords that we want to learn.
Based on this set, we use a spoken query-by-example method to predict which keywords occur in the spoken captions of the speech--image training data.
These predictions are used to automatically construct positive and negative examples for the contrastive loss of the visually grounded speech model.
E.g.\ the 
English speech--image training data might have a caption \word{the boys playing soccer in the park} paired with only a single image.
Using few-shot mining, we can also now pair this utterance with the utterance \word{a dad throws a ball at his boys} as a positive example in the 
loss.
This encourages the model to not only focus on utterances as a whole but to learn within-utterance differences between keywords.
Images that co-occur with mined utterances are similarly used to construct contrasting pairs.

We compare this few-shot mining method to an approach where a visual tagger is used to automatically annotate training images with text labels of words likely occurring in an image~\cite{nortje_towards_2023}.
These generated tags can then be used to sample positive and negative image--caption pairs that contain the same or different keywords, which can again be used in the contrastive loss.
On English data, we show that the few-shot mining approach consistently outperforms this visual tagger scheme in terms of VPKL localisation and detection performance.
We also quantify the drop in performance compared to when transcriptions are used to construct perfect pairs: starting from 50--53\% in the idealised case, detection and localisation F1 drops by roughly 11\%.

We then turn to the actual low-resource language, \yoruba, where we present VPKL results for the first time.
Here we see a larger drop in performance when using few-shot mining compared to using transcriptions. 
This is because the query-by-example matching approach used for mining relies on a self-supervised speech model that is less tailored to \yoruba\ than 
English. 
We also show that it is essential to pretrain the audio branch of the visually grounded speech model on unlabelled \yoruba\ data -- without this, the approach fails, even with perfect pairs.
Qualitative analyses show that, while some scores like precision are modest, the proposed approach provides reasonable outputs on a real low-resource language.
Code will be released upon acceptance.
\section{Visually prompted keyword localisation}
\label{sec:kws}

	

The task of \acf{vpkl} involves two steps:
(i) detecting and (ii) localising a given keyword (specified through an image) in speech utterances.
In the detection step (Fig.~\ref{fig:vpkl}-middle), the model is shown an image query $\boldsymbol{\mathrm{v}}$ depicting a keyword and predicts if the keyword occurs anywhere in a spoken utterance $\boldsymbol{\mathrm{a}}$.
In the localisation step (Fig.~\ref{fig:vpkl}-bottom),
the model predicts the time when the query occurs within the utterance~$\boldsymbol{\mathrm{a}}$.

To perform \ac{vpkl},
we assume we have a dataset of speech and image pairs.
This enables the training of a visually grounded speech model (Sec.~\ref{subsec:multimodalAttention}),
which learns a similarity between images and spoken utterances.
But this is not enough to enable precise detection of specific keywords.
So we further assume access to a small support set of spoken keyword examples.
Based on this set, we automatically mine more training pairs (Sec.~\ref{subsec:sampling}),
which are used for learning to detect the desired keywords.
For localisation, we don't have explicit training data, but we perform it in a weakly-supervised manner by extracting the time frame of the audio that is most similar to the query image.

\subsection{Visually grounded speech model and loss}
\label{subsec:multimodalAttention}

\begin{figure}[!b]
	\centering
	\includegraphics[width=0.85\linewidth]{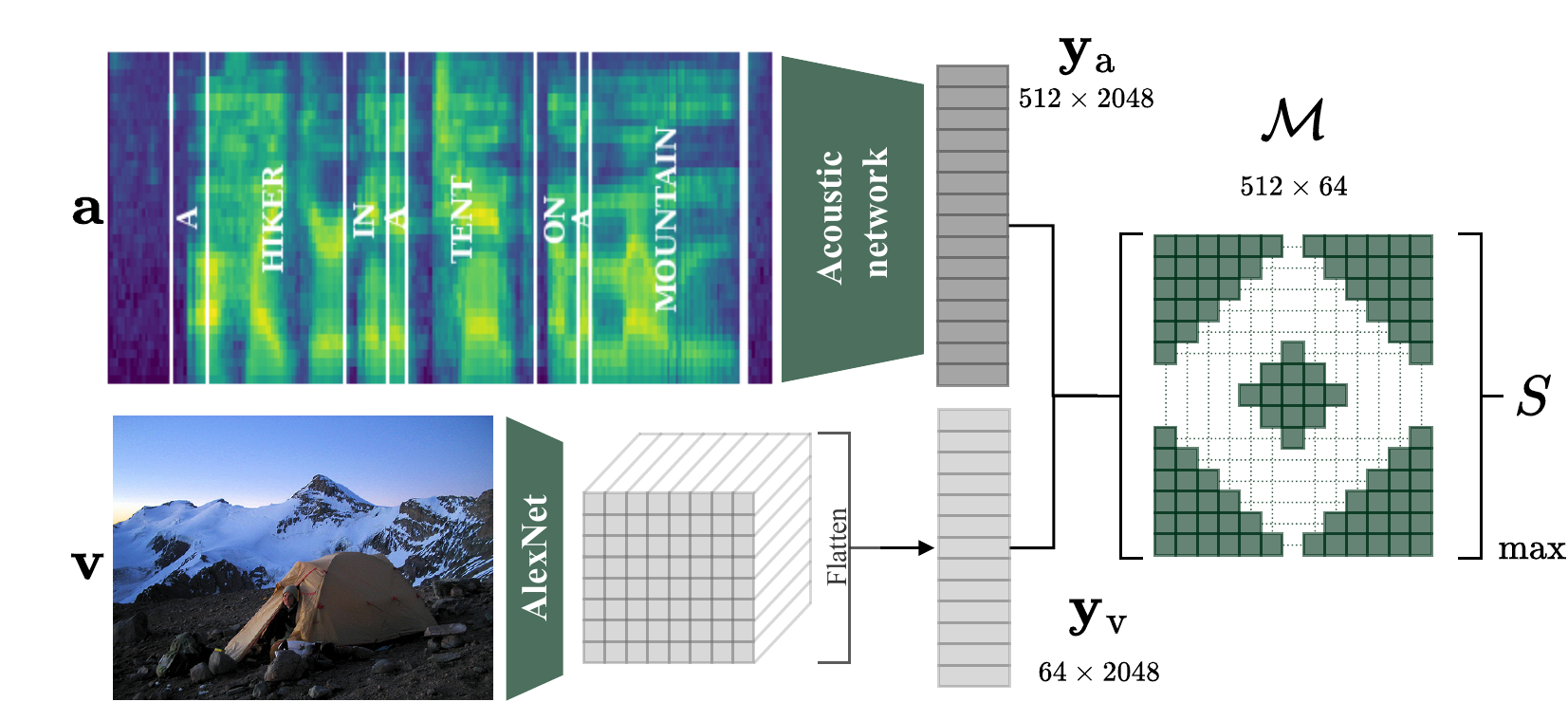}
    \vspace*{-3pt}
	\caption{\locAtt\ consists of a vision and an audio branch connected through a localisation attention mechanism.}
	\label{fig:model}
\end{figure}


The model that we use consists of a vision and an audio branch, connected with an attention mechanism, as shown in Fig.~\ref{fig:model}.
The vision branch is the AlexNet network~\cite{krizhevsky_imagenet_2017} and it encodes
an image $\boldsymbol{\mathrm{v}}$ as a sequence of embeddings $\boldsymbol{\mathrm{y}}_{\textrm{v}}$.
The acoustic branch uses an acoustic network, pretrained on unlabelled speech with contrastive predictive coding (CPC)~\cite{oord_representation_2019}, and it is followed by two BiLSTM layers;
these networks encode a spoken input $\boldsymbol{\mathrm{a}}$ as a sequence of frame embeddings~$\boldsymbol{\mathrm{y}}_{\textrm{a}}$. 
The vision and audio branches are connected by a matchmap attention mechanism~\cite{harwath_jointly_2018} that
computes the dot product between each audio embedding in $\boldsymbol{\mathrm{y}}_{\textrm{a}}$ and each vision embedding in $\boldsymbol{\mathrm{y}}_{\textrm{v}}$,
yielding a similarity matrix $\mathcal{M}$.
To predict at which frames an image query occurs,
we take the maximum over the image axis of $\mathcal{M}$ and obtain a similarity score for each frame.
To get the overall similarity score $S$ for \ac{vpkl} detection, we take the maximum over the entire $\mathcal{M}$.
We refer to this model as \locAtt.
Our model is similar to that of~\cite{nortje_towards_2023}, but the latter employed a much more intricate approach to obtain the similarity score by using context vectors on top of the matchmap, while here we just get the detection score directly.

\locAtt\ is trained as follows.
Paired images and spoken captions in our dataset are used as anchor pairs
($\boldsymbol{\mathrm{a}}$, $\boldsymbol{\mathrm{v}}$).
For each anchor, we sample 
positive utterances $\boldsymbol{\mathrm{a}}_i^+$ and images $\boldsymbol{\mathrm{v}}_i^+$, and negative utterances $\boldsymbol{\mathrm{a}}_i^-$ and images $\boldsymbol{\mathrm{v}}_i^-$.
Positives and negatives are sampled based on a particular keyword.
E.g.\ 
if the keyword is \word{boys}, then the anchor image $\boldsymbol{\mathrm{v}}$ and each positive image $\boldsymbol{\mathrm{v}}_i^+$ contain visual depictions of \image{boys} somewhere in each image;
similarly, on the audio side, the anchor utterance $\boldsymbol{\mathrm{a}}$ and each positive utterance $\boldsymbol{\mathrm{a}}_i^+$ contain \word{boys} somewhere within each utterance.
The visual or spoken representations of \word{boys} do not occur in the negative images $\boldsymbol{\mathrm{v}}_i^-$ or utterances $\boldsymbol{\mathrm{a}}_i^-$.
The idea is that these pairs encourage the model to focus on keywords within utterances and images, rather than focusing on them as a whole.
Based on these pairs, we use a contrastive loss~\cite{nortje_visually_2023-1}:
\vspace*{-3pt}
\begin{equation}
    \scriptsize
    \begin{aligned}	
        \ell = &\ d\left(S(\boldsymbol{\mathrm{a}}, \boldsymbol{\mathrm{v}}), 100\right) + 
        \sum_{i=1}^{N_\textrm{neg}}d\left(S(\boldsymbol{\mathrm{a}}^{-}_{i}, \boldsymbol{\mathrm{v}}), 0\right) + \sum_{i=1}^{N_\textrm{neg}}d\left(S(\boldsymbol{\mathrm{a}}, \boldsymbol{\mathrm{v}}^{-}_{i}), 0\right)\\[-7pt]
        &\qquad+ \sum_{i=1}^{N_\textrm{pos}}d\left(S(\boldsymbol{\mathrm{a}}, \boldsymbol{\mathrm{v}}_{i}^+), 100\right) + \sum_{i=1}^{N_\textrm{pos}}d\left(S(\boldsymbol{\mathrm{a}}_{i}^+, \boldsymbol{\mathrm{v}}), 100\right) \\[-13pt]
        \label{eq:loss}
    \end{aligned}
\end{equation}%
where $d$ is the squared Euclidean distance.
This loss attempts to make the similarity scores $S$ close to 100 for positive pairs while negative pairs are pushed to have scores $S$ close to zero. 

\subsection{Few-shot pair mining}
\label{subsec:sampling}

For a low-resource language, we do not have access to the transcriptions required to sample the positive and negative pairs for~\eqref{eq:loss} above.
To get these pairs, we turn to the few-shot pair mining approach of \cite{nortje_visually_2023-1}.
We start by collecting a small number ($K$) of isolated speech examples for each of the keywords that we want to detect and localise.
These are combined into a support set.
For each keyword, we then use these spoken support set examples
to automatically find utterances containing instances of the same keyword (e.g.\ \word{a group of boys playing soccer} and \word{the boys are climbing a tree}).
Since utterances are also paired with images in our dataset, images can automatically be labelled with the predicted keywords of its paired utterance.
In this way we obtain positive and negative utterances and images for each keyword.
And all this without any transcriptions!

How do we use an example in the support set to find utterances containing the word?
The support set word examples are used as queries in a query-by-example search approach called QbERT.
This method uses HuBERT~\cite{hsu_hubert_2021} to encode speech as a set of discrete units that approximate phones. 
Each query is then scored against each utterance 
in the dataset using a noisy string matching algorithm~\cite{needleman_general_1970}.
We take the mean score across the $K$ word examples per keyword class for each utterance.
The utterances are then ranked from highest to lowest for each keyword and the top $n$ utterances are predicted to contain the keyword.

\begin{table*}[tb]
    \renewcommand{\arraystretch}{1.1}
    \centering
    \newcommand{\ii}[1]{{\footnotesize \textcolor{gray}{#1}}} 
    \newcommand{\deemph}[1]{\textcolor{gray}{#1}}
    \caption{Visually prompted keyword detection and localisation results (\%) on English. Topline models are shown in grey.}
    \vspace*{-7pt}
    \begin{tabularx}{\linewidth}{@{} Cl
        r@{\hspace{0\tabcolsep}}l
        r@{\hspace{0\tabcolsep}}l
        r@{\hspace{0\tabcolsep}}l
        r@{\hspace{0\tabcolsep}}l
        r@{\hspace{0\tabcolsep}}l
        r@{\hspace{0\tabcolsep}}l
        r@{\hspace{0\tabcolsep}}l
        r@{\hspace{0\tabcolsep}}l
        @{}}
        \toprule
        & & 
        \multicolumn{8}{c}{Detection} &
        \multicolumn{8}{c}{Localisation} \\
        \cmidrule(lr){3-10} \cmidrule(l){11-18}
        &
        Model &
        \multicolumn{2}{c}{Precision} &
        \multicolumn{2}{c}{Recall} &
        \multicolumn{2}{c}{F1} &
        \multicolumn{2}{c}{Accuracy} &
        \multicolumn{2}{c}{Precision} &
        \multicolumn{2}{c}{Recall} &
        \multicolumn{2}{c}{F1} &
        \multicolumn{2}{c}{Accuracy} \\
        \midrule 
          \ii{1} & Visually grounded BoW~\cite{olaleye_attention-based_2021} & \textbf{42.29} & & 36.32 & & 39.08 & & 36.63 & & 33.39 & & 31.02 & & 32.17 & & 28.50 & \\
          \ii{2} & Nortje et al.~\cite{nortje_towards_2023} (visual tagger) & 31.02 & & 31.83 & & 31.42 & & 32.02 & & 23.20 & & 25.75 & & 24.21 & & 23.57 & \\
          \ii{3} & \smallma\ (few-shot mined pairs) 
          & 36.94 &  \tiny \deemph{$\pm$2.1} 
          & \textbf{48.80} &  \tiny \deemph{$\pm$1.6} 
          & \textbf{42.03} &  \tiny \deemph{$\pm$1.8} 
          & \textbf{49.16} &  \tiny \deemph{$\pm$1.7} 
          & \textbf{33.72} &  \tiny \deemph{$\pm$1.3} 
          & \textbf{46.52} &  \tiny \deemph{$\pm$1.1} 
          & \textbf{39.09} &  \tiny \deemph{$\pm$1.1} 
          & \textbf{44.21} &  \tiny \deemph{$\pm$0.5}\\
          \addlinespace
          \ii{4} & \textcolor{mygray}{Nortje et al.~\cite{nortje_towards_2023} (ground truth pairs)} & \textcolor{mygray}{48.40} & & \textcolor{mygray} {55.85} & & \textcolor{mygray}{51.86} & & \textcolor{mygray}{56.20} & & \textcolor{mygray}{44.43} & & \textcolor{mygray}{53.79} & & \textcolor{mygray}{48.66} & & \textcolor{mygray}{50.98} & \\
          \ii{5} & \textcolor{mygray}{\smallma\ (ground truth pairs)} 
          & \textcolor{mygray}{63.18} & \tiny \deemph{$\pm$2.1} 
          & \textcolor{mygray}{45.12} & \tiny \deemph{$\pm$3.7} 
          & \textcolor{mygray}{52.62} & \tiny \deemph{$\pm$3.0} 
          & \textcolor{mygray}{45.54} & \tiny \deemph{$\pm$3.6}
          & \textcolor{mygray}{58.98} & \tiny \deemph{$\pm$0.5} 
          & \textcolor{mygray}{43.61} & \tiny \deemph{$\pm$3.1} 
          & \textcolor{mygray}{50.11} & \tiny \deemph{$\pm$2.1} 
          & \textcolor{mygray}{42.84} & \tiny \deemph{$\pm$2.9}\\
          \bottomrule
    \end{tabularx}
    \label{tbl:english_vpkl}
\end{table*}

\begin{figure*}[tb]
\centering
\def\mywidtha{1.5cm}
\def\mywidthb{4.2cm}
\setlength{\tabcolsep}{0.5pt}
\footnotesize
\newcommand{\mylabel}[1]{\sf\scriptsize \color{gray} #1}
\newcommand{\deemph}[1]{\it\color{gray}#1}
\newcommand{\cmark}{\textcolor{green!40!black}{\ding{51}}}%
\newcommand{\xmark}{\textcolor{red!70}{\ding{55}}}%
\begin{tabular}{cc|cc}
  \mylabel{query: \deemph{sand} \deemph{}} &
  \mylabel{query: \deemph{soccer} \deemph{}} &
  \mylabel{query: \deemph{k\d{\`e}k\d{\'e}} (\deemph{bike})} &
  \mylabel{query: \deemph{òkun} (\deemph{ocean})} \\
  \includegraphics[width=\mywidtha, height=\mywidtha]{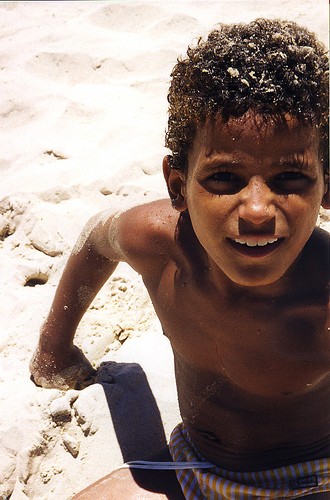} &
  \includegraphics[width=\mywidtha, height=\mywidtha]{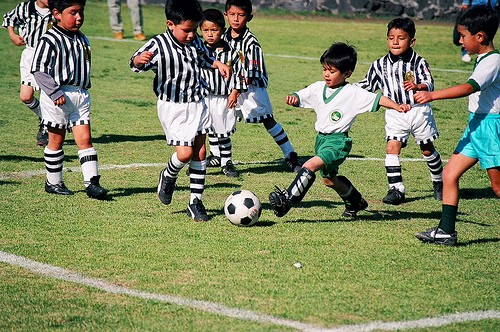} &
  \includegraphics[width=\mywidtha, height=\mywidtha]{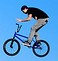} &
  \includegraphics[width=\mywidtha, height=\mywidtha]{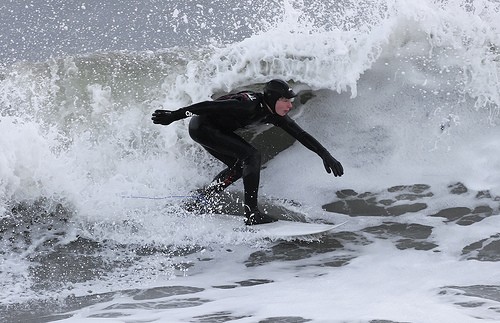} \\
  detection: \cmark · localisation: \cmark &
  detection: \cmark · localisation: \xmark &
  detection: \cmark · localisation: \cmark &
  detection: \xmark · localisation: \xmark \\
  \includegraphics[width=\mywidthb]{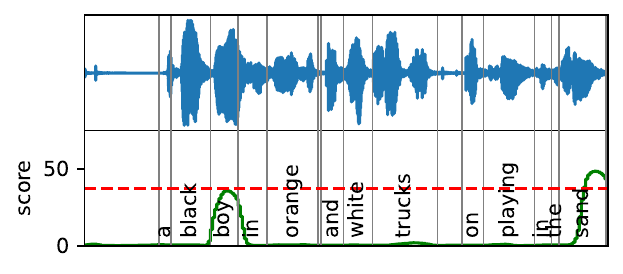} &
  \includegraphics[width=\mywidthb]{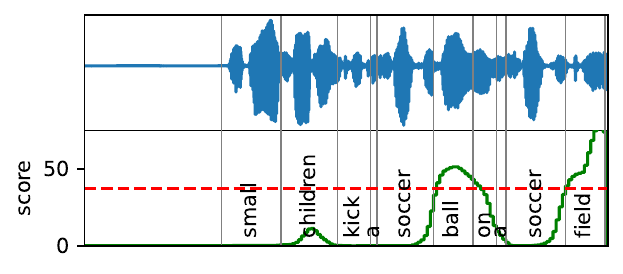} &
  \includegraphics[width=\mywidthb]{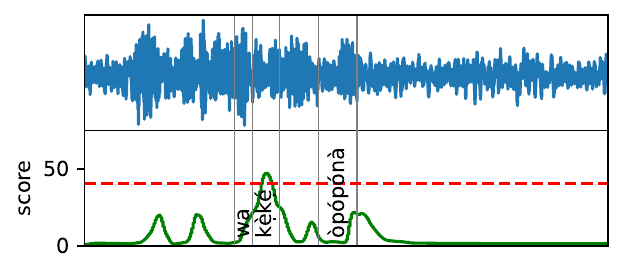} &
  \includegraphics[width=\mywidthb]{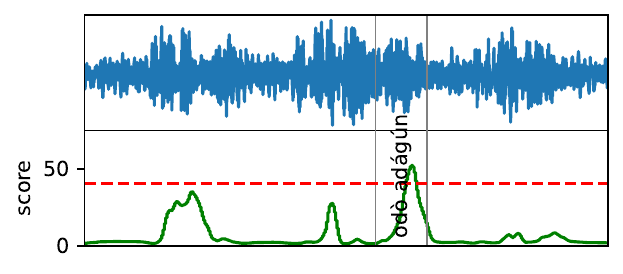} \\
\end{tabular}
\caption{Qualitative samples on English (left) and \yoruba\ (right).
Given a query image, we show the top detected audio sample and the scores for localisation.
We include the corresponding keyword for reference, but this is not seen by the model.
The red dotted line denotes the similarity score threshold $\alpha$,
which determines whether the query image is detected in the audio.
}
\label{examples}
\end{figure*}

\section{English experiments}
\label{sec:english_setup}

To analyse our model and to compare it to previous work, we first perform \ac{vpkl} experiments on English.

\subsection{Experimental setup}
\label{subsec:english_setup}


\hspace{\the\parindent}\textbf{Data.} 
We train an English \locAtt\ on the Flickr8k Audio Captions Corpus (FAAC)~\cite{harwath_deep_2015}, which consists of 8k images each paired with five spoken English captions.
The dataset is split into 30k, 5k and 5k utterances for train, development and test sets.
For the support set, we sample $K = $ 10 examples per keyword from the training and validation sets.
Using forced alignments, we isolate the keywords.
To mine pairs, we use the remainder of the training and development sets as the unlabelled speech dataset and predict that the top $n =$ 200 samples per keyword class contain the keyword.
Utterances are parametrised as mel-spectrograms with a hop length of 10~ms, a window of 25~ms and 40~mel bins. 
These are truncated or zero-padded to 1024 frames.
Images are resized to 224$\times$224 pixels and normalised with means and variances calculated on ImageNet~\cite{deng_imagenet_2009}.

\textbf{Evaluation.}
We follow exactly the same protocol as in~\cite{nortje_towards_2023}.
For each of the 34 keywords, 10 images from the Flickr8k test split were manually cropped to serve as image queries.
The cropped images mostly contain the region corresponding to the keyword, but in some cases also include adjacent objects.
In testing, the similarity score $S$ for an utterance and an image query is calculated.
If $S$ is above a threshold $\alpha$, the keyword depicted in the image query is predicted to be in the utterance.
The $\alpha$ for each model is tuned on the development set.
If a keyword is detected, the frame where the maximum attention score occurs is predicted as the keyword's position. 
We use the ground truth alignments to evaluate the predictions: a true positive is taken when the predicted frame falls within the ground-truth time-span of the keyword. 
It is counted as a mistake if a keyword is falsely detected or the prediction falls outside the time-span.
Each model is trained three times to get mean scores and standard~deviations.


\textbf{Our model.}
The image branch of \locAtt\ is initialised with the convolutional encoder of AlexNet~\cite{krizhevsky_imagenet_2017}, pretrained on ImageNet~\cite{deng_imagenet_2009}.
For the audio branch, we use an acoustic network pretrained using a self-supervised CPC task~\cite{van_niekerk_vector-quantized_2020} on LibriSpeech~\cite{panayotov_librispeech_2015} and the multilingual (English and Hindi) Places dataset~\cite{harwath_vision_2018}.
We take $N_\textrm{neg} = N_\textrm{pos} =$ 4 in~\eqref{eq:loss}, based on development experiments.
The model is trained for 100 epochs using Adam~\cite{kingma_adam_2015}.
A validation task is used for early stopping, with pair mining again used for constructing validation pairs (so transcriptions are never used).


\textbf{Baselines.}
We compare our approach to that of \cite{nortje_towards_2023}.
Instead of pair mining, 
this model uses an external visual tagger to automatically label training images and then use these predicted tags for getting positive and negative pairs in a contrastive loss.
The study~\cite{nortje_towards_2023} also has a topline model that uses transcriptions to get perfect pairs.
We also compare to the visual \ac{BOW} method of~\cite{olaleye_attention-based_2021},
which is queried with written keywords instead of images.
This model is also trained using a visual tagger to
generate textual \ac{BOW} labels for training images. These labels
are then used to train a model that takes speech as input and predicts the location of written keywords as output.
While the task is somewhat different to ours (queries are text instead of images), we can still compare to how well a given keyword is detected and localised.


\subsection{Results}
\label{sec:english_results}

English \ac{vpkl} results are given in Table~\ref{tbl:english_vpkl}.
Line 3 gives the results of our approach.
Compared to the model of \cite{nortje_towards_2023} trained without transcriptions on visual tags (line 2), our new few-shot mining approach is consistently better.
Additionally, our few-shot mining approach outperforms even the unsupervised textual keyword localisation method of \cite{olaleye_attention-based_2021} in line 1.
This is noteworthy given that a written keyword arguably gives a stronger and less variable query signal than an image.
The left part of Fig.~\ref{examples} shows qualitative examples of the few-shot model detecting and localising image queries within utterances.
We see for the keyword \word{soccer}, the system makes a localisation error, but this is reasonable given the ambiguity in the visual query.

To establish the best possible results we could 
get from our approach, we train a \locAtt\ model using ground truth pairs obtained from transcriptions instead of few-shot QbERT-mined pairs (Sec.~\ref{subsec:sampling}). 
By comparing lines 3 and 5, we see that both detection and localisation F1 drop by around 11\%
when moving from the ideal to the sampled pairs (e.g.\ localisation F1 goes from 50.1\% to 39.1\%).
So there is still room for improvement by getting better positive and negative pairs.


We mostly followed the model architecture of~\cite{nortje_towards_2023}, but proposed to simplify the method for getting a similarity score (see Sec.~\ref{subsec:multimodalAttention}).
To see what the influence of the architectural change is, we compare our \locAtt\ topline model (line 5) to the topline of \cite{nortje_towards_2023} (line 4).
While accuracies are somewhat better with the more complex attention mechanism, detection and localisation F1 is better with the simpler model proposed here.

\begin{table*}[t]
    \renewcommand{\arraystretch}{1.1}
    \newcommand{\ii}[1]{{\scriptsize \textcolor{gray}{#1}}} 
    \newcommand{\deemph}[1]{\textcolor{gray}{#1}}
    \centering
    \caption{Keyword detection and localisation results (\%) on \yoruba. Topline models are shown in grey.}
    \vspace*{-7pt}    
    \begin{tabularx}{\linewidth}{@{} Cl
        r@{\hspace{0\tabcolsep}}l
        r@{\hspace{0\tabcolsep}}l
        r@{\hspace{0\tabcolsep}}l
        r@{\hspace{0\tabcolsep}}l
        r@{\hspace{0\tabcolsep}}l
        r@{\hspace{0\tabcolsep}}l
        r@{\hspace{0\tabcolsep}}l
        r@{\hspace{0\tabcolsep}}l
        @{}}
        \toprule
        & & 
        \multicolumn{8}{c}{Detection} &
        \multicolumn{8}{c}{Localisation} \\
        \cmidrule(lr){3-10} \cmidrule(l){11-18}
        &
        Model &
        \multicolumn{2}{c}{Precision} &
        \multicolumn{2}{c}{Recall} &
        \multicolumn{2}{c}{F1} &
        \multicolumn{2}{c}{Accuracy} &
        \multicolumn{2}{c}{Precision} &
        \multicolumn{2}{c}{Recall} &
        \multicolumn{2}{c}{F1} &
        \multicolumn{2}{c}{Accuracy} \\
        \midrule 
        \ii{1} & Visually grounded \ac{BOW}~\cite{olaleye_yfacc_2023} &
        \textbf{38.55} & &
        45.39 & &
        \textbf{41.69} & &
        46.29 & &
        \textbf{22.75} & &
        \textbf{32.89} & &
        \textbf{26.90} & &
        \textbf{26.33} & \\
        \ii{2} & \smallma\ (few-shot mined pairs) &
        7.62           & \tiny $\pm$0.3 &
        \textbf{46.62} & \tiny $\pm$2.1 &
        13.10          & \tiny $\pm$0.4 &
        \textbf{46.62} & \tiny $\pm$2.1 &
         2.43          & \tiny $\pm$0.4 &
        21.60          & \tiny $\pm$1.1 &
         4.36          & \tiny $\pm$0.7 &
        14.73          & \tiny $\pm$1.5 \\
        \addlinespace
        \ii{3} & \textcolor{mygray}{\smallma\ (ground truth pairs \& no CPC)} &
        \textcolor{mygray}{ 8.43} & \tiny \deemph{$\pm$1.1} &
        \textcolor{mygray}{12.00} & \tiny \deemph{$\pm$2.9} &
        \textcolor{mygray}{14.40} & \tiny \deemph{$\pm$1.5} &
        \textcolor{mygray}{50.12} & \tiny \deemph{$\pm$2.9} &
        \textcolor{mygray}{ 2.68} & \tiny \deemph{$\pm$1.0} &
        \textcolor{mygray}{24.74} & \tiny \deemph{$\pm$3.3} &
        \textcolor{mygray}{ 5.08} & \tiny \deemph{$\pm$1.6} &
        \textcolor{mygray}{16.49} & \tiny \deemph{$\pm$3.0} \\
        \ii{4} & \textcolor{mygray}{\smallma\ (ground truth pairs \& CPC)} &
        \textcolor{mygray}{59.96} & \tiny \deemph{$\pm$2.3} &
        \textcolor{mygray}{50.42} & \tiny \deemph{$\pm$1.4} &
        \textcolor{mygray}{54.74} & \tiny \deemph{$\pm$0.2} &
        \textcolor{mygray}{50.42} & \tiny \deemph{$\pm$1.4} &
        \textcolor{mygray}{45.08} & \tiny \deemph{$\pm$2.9} &
        \textcolor{mygray}{43.31} & \tiny \deemph{$\pm$0.7} &
        \textcolor{mygray}{44.13} & \tiny \deemph{$\pm$1.1} &
        \textcolor{mygray}{37.86} & \tiny \deemph{$\pm$0.0} \\
        \bottomrule
    \end{tabularx}
    \label{tbl:yoruba_results}
\end{table*}

\section{Low-resource experiments: \yoruba}
\label{sec:yoruba}

We train a \yoruba\ \locAtt\ to detect and localise an image query depicting a keyword in a \yoruba\ spoken utterance.

\subsection{Experimental setup}
\label{sec:yoruba_setup}

\hspace{\the\parindent}
\textbf{Data.}
For the \yoruba\ experiments, we use the \yoruba\ version of the FAAC dataset, called YFACC~\cite{olaleye_yfacc_2023}.
This is a single-speaker dataset containing a single spoken \yoruba\ caption for each of the 8k Flickr images.
The dataset has 7k, 500 and 500 utterances in its train, development and test sets, respectively.
We manually isolate $K =$~5 spoken examples for each of the 34 keywords from the training and validation sets to obtain the support set.
We use the remainder of the train and development sets as the unlabelled speech dataset for pair mining.
Because this dataset is much more limited than the English case,  to set $n$,
we use the actual number 
of samples in the training and validation sets in which the keyword occurs.


\textbf{Models.}
There are a few changes in the \yoruba\ model compared to the English one (Sec.~\ref{subsec:english_setup}).
First, for pair mining (Sec.~\ref{subsec:sampling}) we replace the English HuBERT in \qbert\ with a multilingual HuBERT trained on English, French and Spanish~\cite{lee_textless_2022}.
The idea is that multilingual representations would be more robust on the unseen language. To tailor the representations to \yoruba\ even more, we train the clustering model
on background \yoruba\ data consisting of 51 hours of Bible recordings \cite{meyer_bibletts_2022, ogayo_building_2022}.
This model gives the discrete units for pair mining.
To initialise the audio branch of the \yoruba\  model, we also use the \yoruba\ Bible data to train the CPC model (Sec.~\ref{subsec:multimodalAttention}). 

\textbf{Evaluation.} 
We use the same image queries for the 34 keywords used in the English task.
The only difference here is that instead of the English utterances, we use the \yoruba\ utterances from the YFACC test set as search utterances.

\subsection{\yoruba\ \ac{vpkl} results}
\label{sec:yoruba_results}

Table~\ref{tbl:yoruba_results} reports the \yoruba\ \ac{vpkl} scores.
Line~2 shows the scores achieved by our \yoruba\ few-shot \locAtt\ model.
This is the first time \ac{vpkl} is performed on an actual low-resource language.
This is also only the second time that keyword localisation is performed on a low-resource language with a visually grounded model, with the first being the 
model 
in line~1 (which takes text queries instead of images).
The detection recall and accuracy scores of the few-shot \locAtt\ (line 2) are competitive to the visual \ac{BOW} model (line 1).
However, the detection precision and localisation scores are 
lower. 
To investigate why this happens, we look at a \yoruba\ \locAtt\ model trained on ground truth pairs (line 4).
This topline model outperforms the \ac{BOW} model with roughly 4--13\% on detection and 10--18\% on localisation. 
In terms of precision,
\locAtt\ outperforms the visual \ac{BOW} model by roughly 21\% on detection and 22\% on localisation.
It therefore seems that the pair mining (Sec.~\ref{subsec:sampling}) is responsible for the poorer scores in line 2, and in particular for the worse precision.
To support this further,
we found that the accuracy of the \yoruba\ mined pairs is 37\% whereas the English mined pairs are 70\%. 
Improving the mined pairs could therefore lead to a very accurate \ac{vpkl} model.
The major difference in the pair mining implementation in \yoruba\
is that the HuBERT model has not been seen any \yoruba\ data. This seems to be crucial for accurate mining.

To further show the importance of the representations being tailored to the target language beforehand, we investigate
the contribution of the \yoruba\ CPC initialisation. 
In line~3, we retrain the ground truth \locAtt\ model 
from a random initialisation without warm-starting from a \yoruba\ CPC model.
Comparing this model to the ground truth model in line~4, we see that CPC initialisation on the target language is essential.
This highlights the advantage of using large unlabelled data to improve low-resource models through self-supervised learning.

The right part of Fig.~\ref{examples} shows qualitative examples of the \yoruba\ few-shot model performing \ac{vpkl}.
In the \word{\ocean}  (\word{ocean}) example, the wrong keyword is detected and localised, \word{\pool} (\word{pool}), which is reasonable given the query.
\section{Discussion}
\label{sec:discussion}

The visually prompted keyword localisation (\ac{vpkl}) task is not a typical mainstream task.
Its formulation involves some nuanced assumptions that are worth further discussion.

\textbf{Visual queries.}
Is \ac{vpkl} really useful in low-resource settings?
E.g.\ query-by-example search could be done using spoken queries rather than images.
Or if we want to search speech in a low-resource language, we could use a \ac{BOW}-based visually grounded speech model~\cite{kamper_visually_2018}, e.g.\ allowing \yoruba\ speech to be searched with English written keywords.
We respond that a visual query is more flexible than either a textual or a spoken query: it can allow a user to search for words that they do not know or, compared to \ac{BOW}-based approaches, to search for words outside of the vocabulary of the visual tagger that is used for supervision.

\textbf{Multiple query objects.}
The \ac{vpkl} task implicitly assumes that the query image refers to a single keyword. 
In our approach we try to achieve this by cropping the most relevant region, but this is not always perfect;
as seen in Fig.~\ref{fig:vpkl}, multiple objects may appear in an image.
An interesting future direction would be to extend the framework to support multi-object settings,
enabling the system to distinguish and localize several keyword referents within the input image.

\textbf{Speech representations.}
Our approach relies on pre-trained HuBERT representations,
which are not optimized for \yoruba\ or other low-resource languages.
Although these features transfer reasonably well, they may miss language-specific acoustic properties,
which are important for keyword localization.
A potential solution is to train a \yoruba-specific HuBERT on unlabeled speech to obtain more tailored representations and improve performance in cross-lingual low-resource retrieval.

\textbf{Few-shot samples.}
A more important limitation of our approach (and one that we agree should be addressed) is that we rely on a few-shot support set containing the keywords we would want to search for.
This makes the approach applicable in low-resource settings, but it means that the vocabulary is constrained.
Future work will look at removing the support set by adapting QbERT to compare whole utterances in a fully unsupervised mining approach, thereby enabling search for arbitrary words.

\section{Conclusions}
\label{sec:conclusions}

We performed \acf{vpkl}~-- detecting and localising an image query depicting a keyword in spoken utterances -- in a low-resource setting.
We did this by building on previous work that followed an idealised scenario on English data.
To make \ac{vpkl} applicable in real low-resource settings, we proposed a few-shot approach to automatically mine positive and negative pairs in a contrastive loss for training a visually grounded speech model.
The few-shot method relies on a small set of isolated examples for the keywords of interest.
Coupled with a simpler attention mechanism than in previous work~\cite{nortje_towards_2023}, we showed that this real low-resource approach is effective in \ac{vpkl} experiments on English and \yoruba.
Future work directions include
adding support for multiple visual queries,
adapting the speech representations to the target language, and
removing the need of few-shot support set.

\bibliographystyle{IEEEtran}
\newpage
\bibliography{mybib}

@inproceedings{harwath_jointly_2018,
	title = {Jointly discovering visual objects and spoken words from raw sensory input},
	booktitle = {Proc. {ECCV}},
	author = {Harwath, David and Recasens, Adria and Suris, Didac and Chuang, Galen and Torralba, Antonio and Glass, James},
	year = {2018},
}

@inproceedings{harwath_deep_2015,
	title = {Deep multimodal semantic embeddings for speech and images},
	booktitle = {Proc. {ASRU}},
	author = {Harwath, David and Glass, James},
	year = {2015},
}

@inproceedings{kingma_adam_2015,
	title = {Adam: {A} method for stochastic optimization},
	booktitle = {Proc. {ICLR}},
	author = {Kingma, Diederik and Ba, Jimmy},
	year = {2015},
}

@article{chrupala_visually_2022,
	title = {Visually grounded models of spoken language: {A} survey of datasets, architectures and evaluation techniques},
	journal = {J. Artif. Intell. Res.},
	author = {Chrupa{\l }a, Grzegorz},
	year = {2022},
}

@inproceedings{kamper_visually_2018,
	title = {Visually grounded cross-lingual keyword spotting in speech},
	booktitle = {Proc. {SLTU}},
	author = {Kamper, Herman and Roth, Michael},
	year = {2018},
}

@inproceedings{peng_fast-slow_2022,
	title = {Fast-slow transformer for visually grounding speech},
	booktitle = {Proc. {ICASSP}},
	author = {Peng, Puyuan and Harwath, David},
	year = {2022},
}

@inproceedings{olaleye_attention-based_2021,
	title = {Attention-based keyword localisation in speech using visual grounding},
	booktitle = {Proc. {Interspeech}},
	author = {Olaleye, Kayode and Kamper, Herman},
	year = {2021},
}

@article{oord_representation_2019,
	title = {Representation learning with contrastive predictive coding},
	journal = {arXiv preprint arXiv:1807.03748},
	author = {Oord, Aaron van den and Li, Yazhe and Vinyals, Oriol},
	year = {2019},
}

@inproceedings{panayotov_librispeech_2015,
	title = {Librispeech: {An} {ASR} corpus based on public domain audio books},
	booktitle = {Proc. {ICASSP}},
	author = {Panayotov, Vassil and Chen, Guoguo and Povey, Daniel and Khudanpur, Sanjeev},
	year = {2015},
}

@inproceedings{harwath_vision_2018,
	title = {Vision as an interlingua: {Learning} multilingual semantic embeddings of untranscribed speech},
	booktitle = {Proc. {ICASSP}},
	author = {Harwath, David and Chuang, Galen and Glass, James},
	year = {2018},
}

@inproceedings{deng_imagenet_2009,
	title = {{ImageNet}: {A} large-scale hierarchical image database},
	booktitle = {Proc. {CVPR}},
	author = {Deng, Jia and Dong, Wei and Socher, Richard and Li, Li-Jia and Li, Kai and Fei-Fei, Li},
	year = {2009},
}

@inproceedings{van_niekerk_vector-quantized_2020,
	title = {Vector-quantized neural networks for acoustic unit discovery in the {ZeroSpeech} 2020 challenge},
	copyright = {All rights reserved},
	booktitle = {Proc. {Interspeech}},
	author = {van Niekerk, Benjamin and Nortje, Leanne and Kamper, Herman},
	year = {2020},
}

@inproceedings{nortje_towards_2023,
	title = {Towards visually prompted keyword localisation for zero-resource spoken languages},
	copyright = {All rights reserved},
	booktitle = {Proc. {SLT}},
	author = {Nortje, Leanne and Kamper, Herman},
	year = {2023},
}

@article{krizhevsky_imagenet_2017,
	title = {{ImageNet} classification with deep convolutional neural networks},
	journal = {ACM},
	author = {Krizhevsky, Alex and Sutskever, Ilya and Hinton, Geoffrey E.},
	year = {2017},
}

@article{hsu_hubert_2021,
	title = {{HuBERT}: {Self}-supervised speech representation learning by masked prediction of hidden units},
	journal = {ACM},
	author = {Hsu, Wei-Ning and Bolte, Benjamin and Tsai, Yao-Hung Hubert and Lakhotia, Kushal and Salakhutdinov, Ruslan and Mohamed, Abdelrahman},
	year = {2021},
}

@article{needleman_general_1970,
	title = {A general method applicable to the search for similarities in the amino acid sequence of two proteins},
	journal = {J. Mol. Biol.},
	author = {Needleman, Saul B. and Wunsch, Christian D.},
	year = {1970},
}

@inproceedings{olaleye_yfacc_2023,
	title = {{YFACC}: {A} {Yor{\`u}b{\'a}} speech-image dataset for cross-lingual keyword localisation through visual grounding},
	booktitle = {Proc. {SLT}},
	author = {Olaleye, Kayode and Oneață, Dan and Kamper, Herman},
	year = {2023},
}

@article{nortje_visually_2023-1,
	title = {Visually grounded few-shot word learning in low-resource settings},
	copyright = {All rights reserved},
	journal = {arXiv preprint arXiv:2306.11371},
	author = {Nortje, Leanne and Oneață, Dan and Kamper, Herman},
	year = {2023},
}

@inproceedings{chrupala_representations_2017,
	title = {Representations of language in a model of visually grounded speech signal},
	booktitle = {Proc. {ACL}},
	author = {Chrupa{\l }a, Grzegorz and Gelderloos, Lieke and Alishahi, Afra},
	year = {2017},
}

@inproceedings{scholten_learning_2021,
	title = {Learning to recognise words using visually grounded speech},
	booktitle = {Proc. {ISCAS}},
	author = {Scholten, Sebastiaan and Merkx, Danny and Scharenborg, Odette},
	year = {2021},
}

@article{scharenborg_speech_2020,
	title = {Speech technology for unwritten languages},
	journal = {IEEE/ACM TASLP},
	author = {Scharenborg, Odette and Besacier, Laurent and Black, Alan and Hasegawa-Johnson, Mark and Metze, Florian and Neubig, Graham and St{\"u}ker, Sebastian and Godard, Pierre and M{\"u}ller, Markus and Ondel, Lucas and Palaskar, Shruti and Arthur, Philip and Ciannella, Francesco and Du, Mingxing and Larsen, Elin and Merkx, Danny and Riad, Rachid and Wang, Liming and Dupoux, Emmanuel},
	year = {2020},
}

@inproceedings{meyer_bibletts_2022,
	title = {{BibleTTS}: {A} large, high-fidelity, multilingual, and uniquely {African} speech corpus},
	booktitle = {Proc. {Interspeech}},
	author = {Meyer, Josh and Adelani, David Ifeoluwa and Casanova, Edresson and {\"O}ktem, Alp and Weber, Daniel Whitenack Julian and Kabongo, Salomon and Salesky, Elizabeth and Orife, Iroro and Leong, Colin and Ogayo, Perez and Emezue, Chris and Mukiibi, Jonathan and Osei, Salomey and Agbolo, Apelete and Akinode, Victor and Opoku, Bernard and Olanrewaju, Samuel and Alabi, Jesujoba and Muhammad, Shamsuddeen},
	year = {2022},
}

@inproceedings{ogayo_building_2022,
	title = {Building {African} voices},
	booktitle = {Proc. {Interspeech}},
	author = {Ogayo, Perez and Neubig, Graham and Black, Alan W.},
	year = {2022},
}

@inproceedings{lee_textless_2022,
	title = {Textless speech-to-speech translation on real data},
	booktitle = {Proc. {NAACL}},
	author = {Lee, Ann and Gong, Hongyu and Duquenne, Paul-Ambroise and Schwenk, Holger and Chen, Peng-Jen and Wang, Changhan and Popuri, Sravya and Adi, Yossi and Pino, Juan and Gu, Jiatao and Hsu, Wei-Ning},
	editor = {Carpuat, Marine and de Marneffe, Marie-Catherine and Meza Ruiz, Ivan Vladimir},
	year = {2022},
}
\end{document}